%% file: ms.tex
\documentclass[10pt,twocolumn,letterpaper]{article}
\usepackage[pagebackref=true,breaklinks=true,letterpaper=true,colorlinks,bookmarks=false]{hyperref}
\usepackage{cvpr}
\usepackage{times}
\usepackage{epsfig}
\usepackage{graphicx}
\usepackage{amsmath}
\usepackage{booktabs}
\usepackage{amssymb}

\usepackage{xcolor}
\usepackage{graphicx}
\usepackage{caption} 
\usepackage{subcaption}
\usepackage[normalem]{ulem}
\usepackage{float}
\usepackage{dblfloatfix}
\usepackage{multirow}
\usepackage[nocompress]{cite}
\usepackage{textcomp}

\def\R{\mathbb{R}}
\newcommand\norm[1]{\left\lVert#1\right\rVert}

\cvprfinalcopy 

\def\cvprPaperID{8531} 

\ifcvprfinal\fi
\begin{document}

\title{OptiBox: Breaking the Limits of Proposals for Visual Grounding}

\author{Zicong Fan\textsuperscript{1}\quad Si Yi Meng\textsuperscript{1}\quad Leonid Sigal\textsuperscript{1, 2, 3}\quad James J. Little\textsuperscript{1}\\
\textsuperscript{1}University of British Columbia\quad \textsuperscript{2}Vector Institute for AI\quad  \textsuperscript{3}Canada CIFAR AI Chair\\
{\tt\small \{fan, mengxixi, lsigal, little\}@cs.ubc.ca}}
\maketitle

\input{body.tex}

{
\small
\bibliographystyle{ieee_fullname}
\bibliography{egbib}
}

\end{document}

%% file: body.tex
\input{sections/abstract.tex}
\input{sections/introduction.tex}

\input{sections/related_works.tex}

\input{sections/method.tex}

\input{sections/experiments.tex}

\input{sections/conclusion.tex}


%% file: sections/abstract.tex
\begin{abstract}
The problem of language grounding has attracted much attention in recent years due to its pivotal role in more general image-lingual high level reasoning tasks (e.g., image captioning, VQA). 
Despite the tremendous progress in visual grounding, the performance of most approaches has been hindered by the quality of bounding box proposals obtained in the early stages of all recent pipelines.
To address this limitation, we propose a general progressive query-guided bounding box refinement architecture (OptiBox) that leverages global image encoding for added context. 
We apply this architecture in the context of the GroundeR model \cite{rohrbach2016grounding}, first introduced in 2016, which has a number of unique and appealing properties, such as the ability to learn in the semi-supervised setting by leveraging cyclic language-reconstruction.
Using GroundeR + OptiBox and a simple semantic language reconstruction loss that we propose, we achieve state-of-the-art grounding performance in the supervised setting on Flickr30k Entities dataset \cite{plummer2015flickr30k}. 
More importantly, we are able to surpass many recent fully supervised models with only 50\% of training data and perform competitively with as low as 3\%.
\end{abstract}

%% file: sections/introduction.tex
\section{Introduction}

\begin{figure}
\centering
\includegraphics[width=0.47\textwidth]{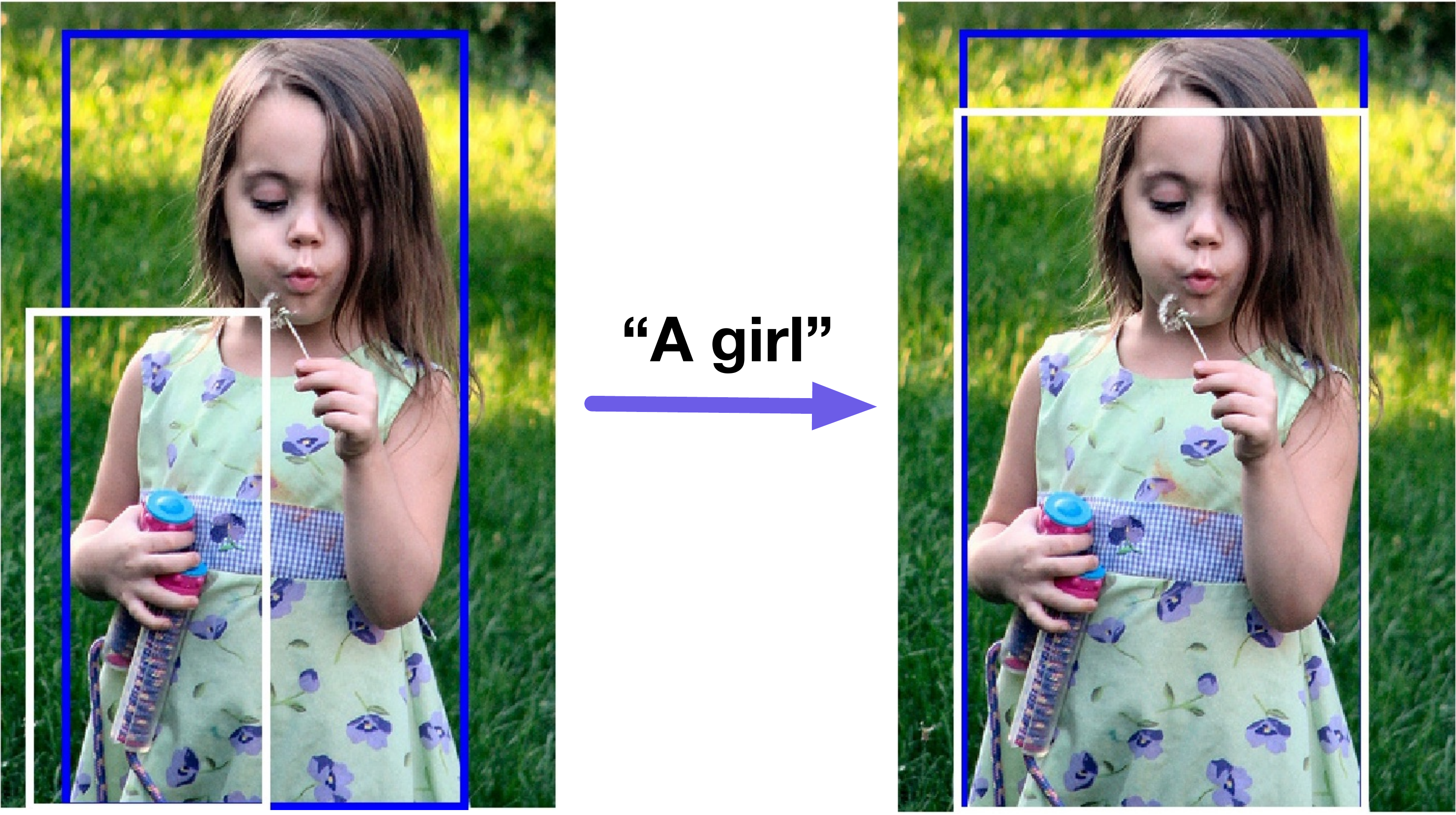}
\caption{Left: the white box represents the raw prediction of a grounding model (GroundeR \cite{rohrbach2016grounding} in this case) for the phrase {\em A girl}. Right: our model (OptiBox) makes appropriate adjustments using query-guided regression, resulting in a much more precise box on the right. The blue box represents the ground-truth.}
\label{fig: cover}
\vspace{-0.05in}
\end{figure}

Visual grounding is the task of associating textual input with corresponding regions in a given image. The problem has attracted much attention in recent years as it plays a vital role in applications such as image captioning \cite{karpathy2015deep} and visual question answering (VQA) \cite{zhu2016visual7w}. Most methods in this field follow a two-stage process \cite{hu2016natural, rohrbach2016grounding, dogan2019neural, plummer2018conditional, mohit2019ground} consisting of \textit{an object proposal stage} that suggests potential bounding boxes a query phrase could ground to, and \textit{a decision stage} that assigns one or more proposed boxes to a query. Despite the various effort to improve visual grounding systems, their performance is bounded by the quality of bounding box proposals in the first stage. We say that a query is correctly grounded if it is assigned a proposal box that is close enough in size and location to the ground-truth annotation box. When all proposals have low overlap with the ground-truth (left of Figure \ref{fig: cover}, for instance), the ability of a grounding model would become extremely limited. 

Although this is an obvious drawback, few methods in visual grounding attempt to address this issue of improving the quality of bounding box proposals.  Chen \etal \cite{chen2017query} introduce a bounding box regression network that leverages reinforcement learning techniques to guide the training. However, their model does not take global visual cues into account, nor is expressive enough for bounding box refinement.
Although Yang \etal \cite{yang2019fast} propose to incorporate query information in the region proposal network to predict bounding boxes that are more strongly correlated with the query. However, they do not refine the predicted boxes. To overcome these limitations, we introduce a bounding box optimization network which we call \mbox{OptiBox} --- that uses global visual cues to refine the predicted proposal boxes progressively and ensures that the box is tight and accurate around the object of interest (Figure \ref{fig: cover} right).

As a proof of concept, we apply OptiBox to the GroundeR model introduced by Rohrbach \etal \cite{rohrbach2016grounding}. Unlike other grounding methods \cite{hu2016natural, dogan2019neural, plummer2018conditional, mohit2019ground, yang2019fast} that have only been shown effective in the fully supervised setting, GroundeR is able to leverage partially labeled or even unlabeled data. When labeled data are available, we suggest a simple semantic reconstruction loss that performs well in both fully-supervised and semi-supervised settings. In particular, with only $\sim 3\%$ annotations, namely, labels for our data, our model outperforms the original fully-supervised GroundeR model ($100\%$ annotations) by a large margin; with $50\%$ annotations, it surpasses most recent fully supervised models; with $100\%$ annotations, it achieves state-of-the-art performance. 
Our ablation studies demonstrate the efficacy of \mbox{OptiBox}, and we expect its application will benefit most other visual grounding frameworks. 

\vspace{0.07in}
\noindent
{\bf Contributions:}
Our contributions are three-fold: 
(1) we propose a general progressive query-guided bounding box refinement architecture that leverages global image encoding for added context;
(2) we apply this architecture in the context of the GroundeR model \cite{rohrbach2016grounding}, illustrating state-of-the-art performance; 
(3) we propose a simple semantic linguistic reconstruction loss which, when coupled with GroundeR, further improves performance in both supervised and, more importantly, semi-supervised settings.
The resulting model can, uniquely, produces close to state-of-the-art (supervised) grounding performance with as little as $3\%$ of the data.

%% file: sections/related_works.tex
\section{Related Work}
\label{sec: vg_attention}
\noindent
\textbf{Visual grounding:} In recent years, there has been considerable progress in visual grounding of phrases. Most authors adopt a two-stage pipeline \cite{rohrbach2016grounding, chen2017query, mohit2019ground, chen2017msrc, hu2016natural,wang2016learning, wang2018learning, plummer2018conditional}: an object proposal stage followed by a grounding decision stage. The object proposal stage generates bounding box proposals from the input image using an object detection model. The grounding decision stage uses the query's linguistic features and the proposal's visual features to score the correspondence between the query and each proposal. The proposal with the highest score is selected as the predicted grounding result. Rohrbach \etal \cite{rohrbach2016grounding} propose the GroundeR model that allows multiple levels of supervision using a reconstruction loss. 
Hu \etal \cite{hu2016natural} use a similar approach based on a caption generation framework but for the supervised setting only. Wang \etal \cite{wang2016learning} apply a deep structure-preserving embedding framework to grounding, which they formulate as a ranking problem. This work is further extended to a similarity network \cite{wang2018learning} and a concept weight branch \cite{plummer2018conditional}. Chen \etal \cite{chen2017query} introduce a query-guided regression framework based on reinforcement learning that refines the proposed boxes using query heuristic in order to break the bottleneck of bounding box proposals. Furthermore, Chen \etal \cite{chen2017msrc} take the contextual information of the phrase into account by penalizing a joint loss for all phrases in a sentence, whereas Dogan \etal \cite{dogan2019neural} encode the queries and proposal features using two independent LSTM modules. 
Bajaj \etal \cite{mohit2019ground} construct a visual graph and a phrase graph to model the pairwise relationship between entities via graph convolutional operations. The convolved visual and linguistic features are merged and refined by a fusion graph in the end. There have also been techniques focusing on single-stage grounding. Xiao \etal \cite{xiao2017weakly} perform weakly-supervised pixel-level grounding with a spatial attention mask generated from the hierarchical structure of the parse tree from the phrase query. Very recent works propose to embed query information in the region proposal stage \cite{yang2019fast,sadhu2019zero}. In contrast, our proposed method shows that with simple architecture, we can learn from features in multiple modalities to obtain a more refined, accurate result in various learning settings.

\vspace{0.07in}
\noindent
\textbf{Bounding box regression:} Bounding box regression has been commonly used in the final stage of an object detector to improve its accuracy. The R-CNN line of work \cite{girshick2014rich, girshick2015fast, ren2015faster,he2017mask} applies a linear regression layer to refine bounding boxes selected from the list of proposals. The model in \cite{gidaris2015object} iteratively refines and merges proposals using a deep CNN regression model. Lin \etal \cite{lin2017focal} use a class-agnostic, convolutional bounding box regressor to correct the offset between an anchor and its closest ground-truth. Additionally, Jiant \etal \cite{jiang2018acquisition} learn to predict the IoU between the proposals and their ground-truth (IoU-Net).
Instead of using a fixed architecture to perform bounding box regression, Rajaram \etal \cite{rajaram2016refinenet} implement an iterative refinement algorithm (RefineNet) that can be trained in a similar fashion as the Faster R-CNN \cite{ren2015faster} network. Roh and Lee \cite{roh2017refining} add a refinement layer for both the object classification and bounding box regression losses. Recently, Rezatofighi \etal \cite{rezatofighi2019generalized} introduce a generalized IoU metric that is more robust to non-overlapping objects, while He \etal \cite{he2019bounding} use the KL divergence between the predicted and ground-truth distribution as the penalty. 
In the task of object segmentation, Pinheiro \etal \cite{pinheiro2016learning} propose SharpMask which augments a traditional feedforward net with a refinement module to produce more accurate segmentation. The unique challenge of incorporating these bounding box regression methods into visual grounding is the multimodality of the task -- we need to ensure the correction is guided by the query, rather than treating it as an object detection refinement procedure.

%% file: sections/method.tex

\section{Approach}

Our grounding model consists of two parts: a grounding module and a box refinement module. Given an image and a query, the grounding module returns a bounding box $\mathbf{r}$ with the highest confidence value. The box refinement module then refines $\mathbf{r}$ and outputs an offset $\mathbf{t'}$ to suggest an adjustment on $\mathbf{r}$. Finally, using the predicted offset $\mathbf{t'}$, we obtain the refined bounding box $\mathbf{\hat{g}}$. 

\subsection{Grounding}
\label{sec: grounder_method}
We adopt the GroundeR model introduced in \cite{rohrbach2016grounding} as our grounding module. An input image first goes through an object detector to obtain $N$ bounding box proposals $\mathbf{r}_i = [r_x, r_y, r_w, r_h]$ of potential objects, where $i\in[1,\dots,N]$. For each proposal $\mathbf{r}_i$, we extract its visual features $\mathbf{x}_i$ from the detection backbone. For the query phrase, we first encode each word using pre-trained word embeddings, which are then passed into an LSTM \cite{hochreiter1997long} to obtain the last hidden state linguistic features $\mathbf{h}$. To combine the two modalities, we first project $\mathbf{x}_i$ and $\mathbf{h}$ both onto a common dimensionality space using two separate fully-connected layers with the ReLU activation. The projected query feature $\mathbf{h}' $ is then added to each of the projected proposal features $\mathbf{x}_i' $ to produce a feature vector $\mathbf{z}_i$ for each proposal. At this point,  $\mathbf{z}_i $ should contain information from both the query and the object $i$. At inference time, we score the correspondence between the pairs by projecting $\mathbf{z}_i$ to a scalar $\alpha_i$ and assign the query to the proposal with the maximum $\alpha_i$. At training time when labeled data are available, the cross-entropy loss $L_{cls}$ is applied against the target box, which is the proposal with the highest IoU out of all with an IoU of at least 0.5 with the ground-truth.

\begin{figure}[t]
    \centering\includegraphics[width=1\linewidth]{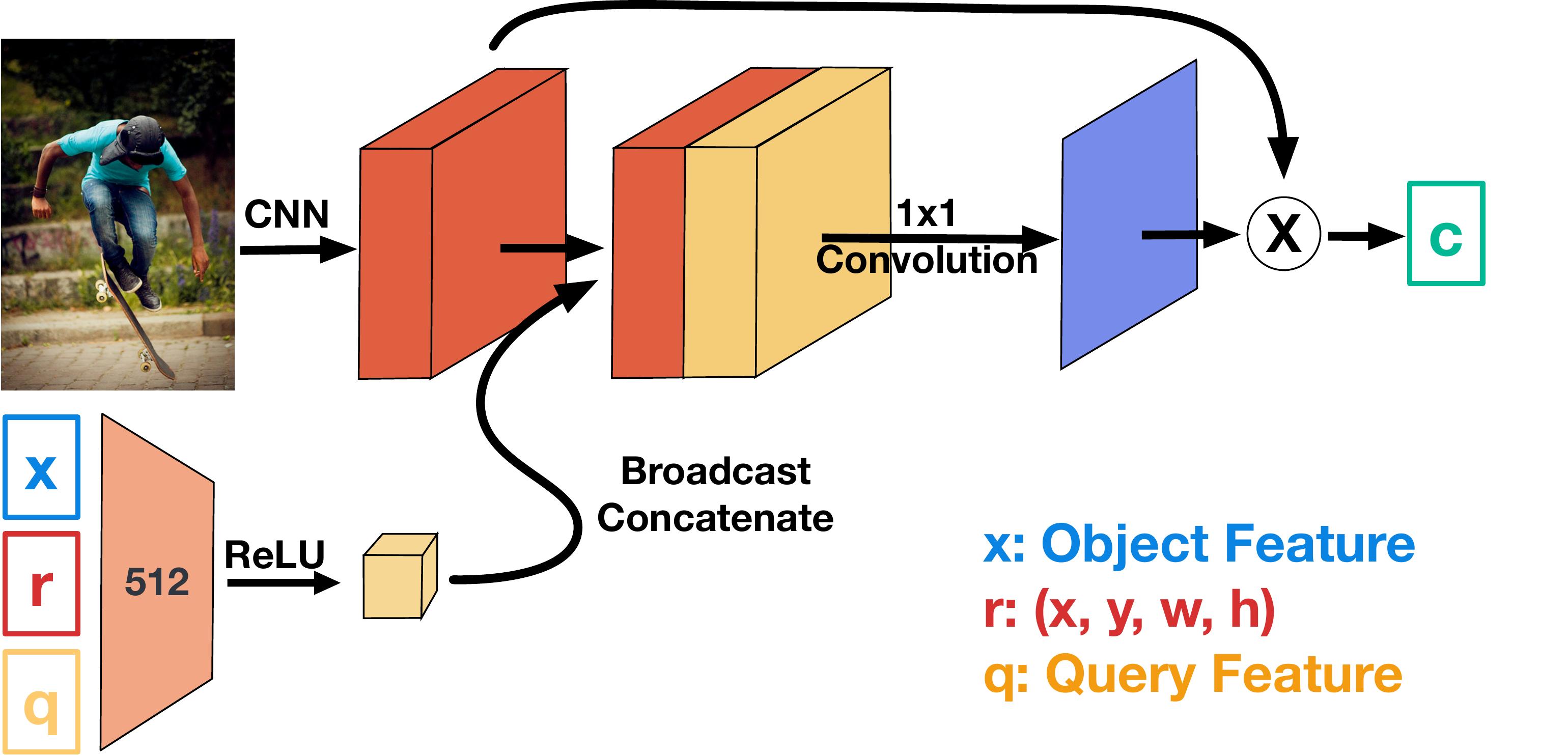}
    \caption{{\bf Our global attention module}. The model first extracts the feature map from the input image using the detection backbone. It then concatenates a box $\mathbf{r}$, the visual feature $\mathbf{x}$ of $\mathbf{r}$, and the LSTM encoded query feature $\mathbf{h}$. The concatenated vector is then projected onto $\mathbb{R}^{512}$. Each spatial grid of the feature map then concatenates with a copy of this $512$-dimensional vector, which results in an attention feature map after a $1\times 1$ convolution. After softmax normalization, the attention map entries are used as the weights to average across all spatial grid locations of the feature map to produce a context vector $\mathbf{c}$. Best viewed in color.}
  \label{fig: global_attn}
\end{figure}

\textbf{Semantic Reconstruction:} In the semi-supervised setting, the original GroundeR \cite{rohrbach2016grounding} model exploits unlabeled data through a phrase reconstruction process: if the model is highly confident about a subset of the proposals, one should be able to reconstruct the phrase solely using the visual features of those boxes. In particular, it normalizes $\alpha_i$ with a softmax function so that $\sum_{i=1}^{N} \alpha_i' = 1$. It then takes a weighted average over the visual features of all proposals,
\begin{equation}
\mathbf{x}_{att}=\sum_{i=1}^{N} \alpha_{i}' \mathbf{x}_{i}, 
\label{eq: convex}
\end{equation}
and learns an LSTM decoder to unroll the original phrase query. A cross-entropy loss is applied to the phrase reconstruction at each time step of the LSTM. Although this seems to be the most natural approach to take, there is an inherent flaw: \emph{Even if the reconstructed phrase is semantically equivalent to the input phrase, it could be penalized if it does not match the input phrase word for word}. For example, ``a little boy'' and ``a young boy'' could refer to the same bounding box, and thus this minor deviation should not be penalized. In other words, the model should strive to reconstruct the semantic meaning of the phrase based on the visual features, rather than trying to match exactly.
To this end, we propose to learn a function $f_{proj}$ that projects the visual feature $\mathbf{x}_{att}$ to the original query semantic space,

\begin{equation}
\hat{\mathbf{h}} =f_{proj}\left(\mathbf{x}_{a t t}\right)=W_{proj} \mathbf{x}_{a t t}+b_{proj}
\end{equation}

and penalize with the semantic loss
\begin{equation}
\mathcal{L}_{sem}=\norm{\mathbf{h}-\hat{\mathbf{h}}}
\end{equation}
in the linguistic latent space. To balance the classification loss and the semantic reconstruction loss, we use a hyper-parameter $\lambda$ and mimimize the sum of the two losses:
\begin{equation}
\mathcal{L}=\lambda \mathcal{L}_{cls}+\mathcal{L}_{sem}.
\label{eq: joint_loss}
\end{equation}

\begin{figure}[t]
    \centering\includegraphics[width=1\linewidth]{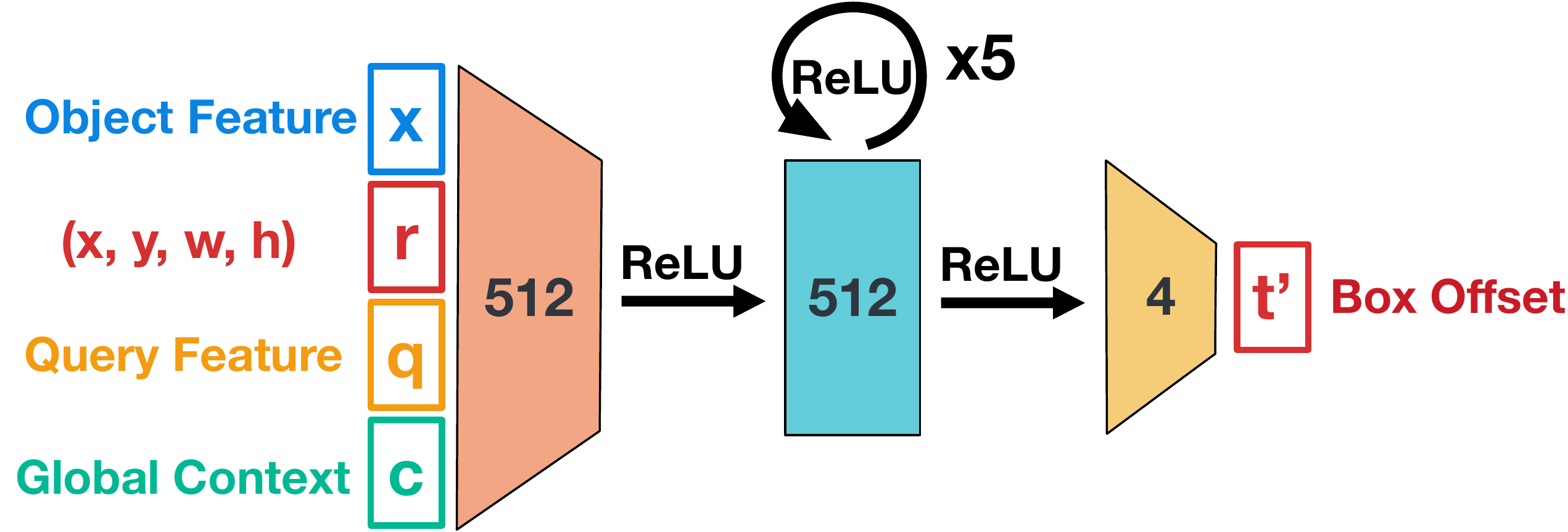}
  \caption{\textbf{OptiBox: our proposed bounding box refinement model}. Given a predicted box from a grounding network, we concatenate the visual features $\mathbf{x}$ corresponding to the box, the bounding box $\mathbf{r}$ itself, the query's linguistic features $\mathbf{q}$, and the global context features $\mathbf{c}$ and project using feedforward layers. The final projection yields a $4$-dimensional bounding box offset. The numbers on the layers indicate the corresponding layer output dimensions. Best viewed in color.}
  \label{fig: regression}
\end{figure}

\subsection{OptiBox: A Bounding Box Refinement Model}

In two-staged visual grounding, the first stage is often crucial as the quality of the proposals fundamentally limits the latter stage. For instance, if all proposals generated significantly offset from objects in the image, the grounding module would not be able to learn and predict well due to the lack of suitable candidates. Even in the case when the query is grounded correctly, the selected bounding box may deviate by a noticeable amount from the actual object indicated by the query, since the grounding module has not been optimized to perform further adjustments beyond its predictions. For these reasons, we propose to use a bounding box regression module to refine the proposals. Note that our method is essentially metric-agnostic: it will yield tighter and more accurate bounding boxes for visual grounding regardless of the evaluation metric used. 

For convenience, we drop the subscript $i$ for the predicted bounding box. OptiBox takes advantage of several available components of our method: the bounding box $\mathbf{r}$, the visual feature $\mathbf{x}$, the original query feature $\mathbf{q}$, and the global feature map $G$ of the entire image. Figure \ref{fig: regression} depicts the simple refinement architecture of OptiBox. We first encode the original query into a language feature vector $\mathbf{q}$ using a separate LSTM. We then apply global attention to pool context information from the image to obtain a global context vector $\mathbf{c}$ (which we will discuss in detail shortly). We concatenate $\mathbf{x}, \mathbf{r}, \mathbf{q}, \mathbf{c}$ into a single vector and project it to a lower-dimensional space for refinement. The refinement process consists of 5 weight-sharing fully-connected layers of the same size. Empirically, using additional layers only yield marginal returns, and the choice of weight sharing is to avoid overfitting.
Finally, we project the output of the final refinement layer to a box offset vector $\mathbf{t}' \in \R^4$, and all fully connected layers are followed by the ReLU activation.

Figure \ref{fig: global_attn} illustrates the structure of the aforementioned global attention module. For the input image in question, we extract its feature map from the detection backbone. We perform adaptive average pooling to obtain a feature map $G$. The visual feature $\mathbf{x}$, bounding box $\mathbf{r}$, and the query feature $\mathbf{q}$ are then concatenated and projected to a lower-dimensional vector. Each spatial grid of the feature map receives one copy of this projected vector, yielding a more informative feature map. We then apply a $1\times 1$ convolution to the feature map and obtain an attention map. Finally, we normalize the attention map using softmax and perform a weighted average across all channels of $G$, giving us a global attention vector $\mathbf{c} \in \R^{1024}$. 

To be consistent with widely-used methods in bounding box regression \cite{girshick2014rich}, we adopt the following bounding box offset representation: given a box to regress $\mathbf{r} = [r_x, r_y, r_w, r_h] \in \R^4$ and the groundtruth box $\mathbf{g} = [g_x, g_y, g_w, g_h] \in \R^4$, the targets of our bounding box regression model are defined by
\begin{align}
 \mathbf{t}_{x} &=\left(\mathbf{g}_{x}-\mathbf{r}_{x}\right) / \mathbf{r}_{w}          &  \mathbf{t}_{y} &=\left(\mathbf{g}_{y}-\mathbf{r}_{y}\right) / \mathbf{r}_{h} \nonumber\\
 \mathbf{t}_{w} &=\log \left(\mathbf{g}_{w} / \mathbf{r}_{w}\right)          & \mathbf{t}_{h} &=\log \left(\mathbf{g}_{h} / \mathbf{r}_{h}\right).
 \label{eq: offset1}
\end{align}
During inference time, given the chosen proposal box $\mathbf{r}$ and the predictions from box regression network $\mathbf{t}'$, the regressed box $\mathbf{g}'$ can be obtained by
\begin{align}
\hat{\mathbf{g}}_{x}&=\mathbf{r}_{w}  \mathbf{t}'_x+\mathbf{r}_{x}         &  \hat{\mathbf{g}}_{y}&=\mathbf{r}_{h} \mathbf{t}'_y+\mathbf{r}_{y} \nonumber\\
\hat{\mathbf{g}}_{w}&=\mathbf{r}_{w} \exp \left(\mathbf{t}'_w\right)          & \hat{\mathbf{g}}_{h}&=\mathbf{r}_{h} \exp \left(\mathbf{t}'_h\right).
\label{eq: offset2}
\end{align}

%% file: sections/experiments.tex
\section{Experiments and Results}
\subsection{Implementation Setup}

\begin{table*}[t]
\centering
\begin{tabular}{lcccc} 
\toprule
\multicolumn{1}{c}{\textbf{Approach}} & \textbf{Visual Features} & \textbf{Fine-tune Dataset} & \textbf{\# Proposals} & \textbf{Acc. (\%)}  \\ 
\hline
\textbf{Fully-supervised} \\
SCRC \cite{hu2016natural}   & VGG16 & ImageNet & 100 & 27.80 \\
DSPE \cite{wang2016learning}   & VGG19 & Pascal & 100 & 43.89 \\
GroundeR  \cite{rohrbach2016grounding} & VGG16 & Pascal & 100 & 47.81 \\
CCA \cite{plummer2015flickr30k} & VGG19 & Pascal & 200 & 50.89 \\
MCB $+$ Reg $+$ Spatial \cite{chen2017msrc} & VGG16 & Pascal & 100 & 51.01 \\
Similarity Net \cite{wang2018learning} & VGG19 & Pascal & 200 & 51.05\\
MNN $+$ Reg $+$ Spatial \cite{chen2017msrc} & VGG16 & Pascal & 100 & 55.99 \\
SeqGROUND \cite{dogan2019neural} & VGG16 & N/A & N/A & 61.60 \\
CITE-Resnet \cite{plummer2018conditional}& ResNet101 & COCO & 200 & 61.33 \\
CITE-Pascal\cite{plummer2018conditional}& VGG16 & Pascal & 500 & 59.27     \\
CITE-Flickr30K\cite{plummer2018conditional}& VGG16 & Flickr30K & 500 & 61.89 \\
QRC Net \cite{chen2017query}& VGG16 & Flickr30K & 100 & 65.14 \\
GraphGround - PhraseGraph \cite{mohit2019ground}& VGG16 & Visual Genome & 50 & 60.80 \\
GraphGround \cite{mohit2019ground}& VGG16 & Visual Genome & 50 & 63.87 \\

GraphGround++ \cite{mohit2019ground}& VGG16 & Visual Genome & 50 & 66.93 \\
\hline
One-Stage-Bert \cite{yang2019fast}& Darknet53-FPN & COCO (Flickr30K) & 4032 & 68.69 \\
\hline
\textbf{Ours: Fully-supervised}  \\
GroundeR & ResNet101 & Visual Genome & 50 & 67.04 \\
\textbf{Ours: Semi-supervised}  \\
Annotation \%: 3.12\%  & ResNet101 & Visual Genome & 50 & 58.55 \\
Annotation \%: \hspace{1.7mm} 50\%   & ResNet101 & Visual Genome & 50 & 65.85 \\
\bottomrule
\end{tabular}
\caption{Test accuracy comparison of the state-of-the-art models on the Flickr30k Entities dataset. For our approaches, we use GroundeR with OptiBox in all cases with the semantic loss.}
\label{tab:supervised}
\end{table*}

\noindent
\textbf{GroundeR:} To encode the query phrase, we use 200-dimensional GloVe \cite{pennington2014glove} embeddings pre-trained on the Twitter corpus. The word embeddings are then passed into a single-layer, uni-directional LSTM with hidden size 512. The hidden states are batch-normalized and projected to 128 dimensions. We adapt the ResNet101 network \cite{he2016deep} pre-trained on Visual Genome \cite{krishna2017visual} with the top 200 most frequent object class labels. To be consistent with most existing approaches, we do not finetune the detection backbone on Flickr30k \cite{plummer2015flickr30k}, which is our train and evaluation dataset for grounding. We allow the region proposal network to generate $50$ proposals for each image. We extract the C4 layer of ResNet101 as our global feature map $G$ and perform global average pooling after the ROI head to obtain a 2048-dimensional feature vector $\mathbf{x}_i$ for each proposed region $r_i$. The bounding box visual features $\mathbf{x}_i$ are also batch-normalized, followed by a projection to $\mathbb{R}^{128}$. The resulting vector is summed with the projected hidden states from the query phrase. The aggregated features then go through a fully connected layer that forms the attention over the proposed bounding boxes, and the attention values are penalized against the target using cross-entropy. We train our grounding model for 25 epochs with weight decay $0.0005$. We use the Adam optimizer \cite{kingma2014adam} with batch size 128 and a scheduled learning rate (that decays from 0.001 by $1/10$ at epoch 15 and 25), and we select the model maximizing validation accuracy for reporting on the test set.

\begin{table}[t]
\centering
\begin{tabular}{l|lll} 
\toprule
\textbf{Hyperparam.\textbackslash{}Annot.} & \textbf{3.12}\% & \textbf{50}\%   & \textbf{100}\%  \\ 
\hline
\textbf{Weight decay}                       & 0.01   & 0.0005 & 0.01   \\
\textbf{Semantic loss reg.\,}($\mathbf{\lambda}$)         & 10     & 100    & 100    \\
\bottomrule
\end{tabular}
\caption{Grid search results for key hyperparameters.}
\label{tab:grid-search}
\end{table}

In the semi-supervised setting, we optimize the classification loss and the semantic reconstruction loss jointly (see Eq. \ref{eq: joint_loss}). The semantic reconstruction loss uses the $\ell_1$-distance for its empirical performance. Note that the value of weight decay and the choice of $\lambda$ can significantly impact the accuracy. Thus we select them from the validation set via a grid search. The selected values are in Table \ref{tab:grid-search}.

\vspace{0.07in}
\noindent
{\bf OptiBox}: To be consistent with the grounding model, we also encode each word of the original query by the 200-dimensional GloVe embeddings from the Twitter corpus and feed each word embedding sequentially through a separate, uni-directional LSTM with hidden dimension 512. The query feature $\mathbf{q}$ is then extracted from the final hidden state of the LSTM. As mentioned, our box regression model makes use of the query feature $\mathbf{q}$ with $512$ dimension, the visual feature $\mathbf{x}$ with $2048$ dimensions, the box with $4$ dimensions, and the global feature map with dimensions $1024\times 10 \times 10$. The shared dimension of refinement layers in Figure \ref{fig: regression} are in 512 dimensions; the fully-connected layer to project $[\mathbf{x}; \mathbf{r}; \mathbf{q}]$ before concatenating with each channel of feature map $G$ is 512 dimensions. The ReLU activation is added between fully-connected layers of Figure \ref{fig: regression}. It is also used before broadcasting the local features onto the feature map $G$. Since the performance of the bounding box regression is conditioned on the outputs of the grounding model, we begin training the regression model when the grounding model converges. To train the regression model, we use Adam \cite{kingma2014adam} with a learning rate of $0.0001$ and a batch size of 128 until convergence. Again, the $\ell_1$-loss works the best in our validation. Thus, we apply the $\ell_1$-loss between the predicted bounding box offset $\mathbf{t}' \in \R^4$ and the target offset $\mathbf{t}$ for supervision (see Equations \ref{eq: offset1} and \ref{eq: offset2}). 

In order to obtain highly informative linguistic features, we pre-train an LSTM autoencoder using queries in the training set. The autoencoder is trained using Adam \cite{kingma2014adam} for $60$ epochs with an initial learning rate of $0.0001$ on batches of size $128$. The learning rate decays by a factor of $0.1$ at epoch $40$. Once the autoencoder converges, we use it to initialize the weights of the LSTM encoders in GroundeR and OptiBox. Previous methods \cite{dogan2019neural, mohit2019ground, yang2019fast, wang2016learning} have shown that learning an image projection and a query projection with a ranking loss \cite{rankingloss} to ensure the projected vectors in a common space helps. Therefore, with the LSTM encoder initialized, we freeze its weights and pre-train the two linear layers mentioned in Section \ref{sec: grounder_method} to project the image feature $\mathbf{x_i}$ to $\mathbf{x_i'}$ and to project the query feature $\mathbf{q}$ to $\mathbf{q'}$ using the ranking loss. We trained the projections with Adam with a learning rate of $0.0001$ and a batch size of $256$ for $20$ epochs. The learning rate decays by a factor of $0.1$ at epochs $10$, $15$, and $18$.

\subsection{Dataset and Evaluation}

\begin{table}[t]
\begin{tabular}{ccc}
\toprule
\textbf{Approach}   & \textbf{\small{Prop. UB (\%)}} & \textbf{\small{Acc. (\%)}} \\
\hline
GroundeR \cite{rohrbach2016grounding}& 77.90                              & 47.81                  \\
RPN+QRN \cite{chen2017query}& 71.25                              & 53.48                  \\
SS+QRN \cite{chen2017query}& 77.90                              & 55.99                  \\
PGN+QRN \cite{chen2017query}& 89.61                              & 60.21                  \\
One-Stage-Bert \cite{yang2019fast}& 95.48                              & 68.69                  \\
\hline
GroundeR & 84.00                              & 62.15                  \\
GroundeR + OptiBox & 84.00                              & 65.20                \\
GroundeR + SL & 84.00                              & 62.25                  \\
GroundeR + SL + OptiBox& 84.00                              & 67.04                 \\
\bottomrule
\end{tabular}
\caption{Comparison of proposal upper bounds of various state-of-the-art models with ours (last four). SL indicates using semantic loss. In the QRN network, RPN, SS and PGN are different proposal methods.}
\label{tab:proposal_ub}
\end{table}

\begin{table}[t]
\centering
\begin{tabular}{lclcl} 
\toprule
\multicolumn{1}{c}{\textbf{Approach}} & \textbf{Feature} & \textbf{Acc. (\%)}\\ 
\hline
GroundeR \cite{rohrbach2016grounding} & VGG16 CLS & 41.56 \\
GroundeR \cite{rohrbach2016grounding} & VGG16 DET & 47.81\\
\hline
GroundeR & ResNet101 DET & 62.15 \\
GroundeR + OptiBox & ResNet101 DET & 65.20 \\
\bottomrule
\end{tabular}
\caption{Comparison of the original GroundeR model and our models without semantic loss. CLS and DET denote the model is optimized for classification and detection task, respectively. The median IoU of model prediction for GroundeR with ResNet101 feature is 0.6008 and 0.6617 when OptiBox is added.}
\label{tab: regression_ground}
\end{table}

We evaluate the test time performance of our model on the Flickr30k Entities dataset \cite{plummer2015flickr30k}. The dataset contains 31,783 annotated images in total. Each image has five sentences that contain phrases describing entities in the image. Each phrase is also annotated with the ground-truth bounding boxes within the image. We use the same data split as released by the dataset authors, which consists of 1000 images for validation, 1000 images for testing, and the rest for training. At evaluation time, a predicted bounding box that has a higher than 0.5 IoU with the ground-truth bounding box is considered correctly grounded, and we report the accuracy on the test set, which is the percentage of the test set phrases that are correctly grounded. 

Table \ref{tab:supervised} shows the accuracy of various state-of-the-art models comparing to ours along with the visual features to use, on what dataset the detection backbone is fine-tuned on, and the number of proposed bounding boxes used. We can see that with a simple replacement of the detection backbone in the GroundeR model, we can already achieve a competitive performance against the most recent models. Among these methods, only the QRC Net performs bounding box regression, but does not use a recurrent, shared-weight component. Table \ref{tab:proposal_ub} shows that their model also uses more proposals with a higher proposal upper bound (possibly due to fine-tuning on Flickr30k\cite{plummer2015flickr30k}), yet we are able to exceed their accuracy using only half the number. Comparing to \cite{mohit2019ground}, although this model makes grounding decisions jointly by considering all phrases within a sentence altogether, our accuracy is slightly higher than all variations of it. We recently became aware of a concurrent work called the One-Stage-Bert model \cite{yang2019fast}. Although they have higher accuracy, their detection backbone was again fine-tuned on Flickr30k\cite{plummer2015flickr30k} and implicitly used more proposals through the YOLO framework to achieve a high proposal upper bound. Although we could also have fine-tuned our detection backbone to obtain better results, we avoided doing so to be consistent with the visual grounding literature.

\textbf{Fully Supervised:} Note that the original GroundeR \cite{rohrbach2016grounding} uses VGG16\cite{simonyan2014very} detection features while our version uses ResNet101 detection features.  Results from the CITE model \cite{plummer2018conditional} in Table \ref{tab:supervised} shows that the use of ResNet101 also has some positive impact on grounding performance. 
Table \ref{tab: regression_ground} shows the performance of our model compared to the original GroundeR model in the supervised setting. Indeed, with our enhanced version of GroundeR (before adding OptiBox), we achieve an over $10\%$ improvement in the test accuracy, possibly due to the higher proposal upper bound and better proposal quality. The QRN variants in Table \ref{tab:proposal_ub} shows an example where this could happen. Recall that the proposal upper bound is defined as the proportion of queries with at least one proposal having a $\geq 0.5$ IoU with its ground-truth. Interestingly, looking at Table \ref{tab:proposal_ub}, with the reconstruction loss, when using all annotations, our model performs slightly better than the model without it. We posit that the semantic reconstruction loss acts as a regularizer when annotation is abundant.

\begin{table}[b]
\begin{tabular}{cccc}
\toprule
\textbf{Models \textbackslash Annot. (\%)} & \textbf{3.12\%} & \textbf{50\%}  & \textbf{100\%} \\
\hline
GroundeR \cite{rohrbach2016grounding} & 28.94           & 46.65          & 48.38          \\
\hline
GroundeR + SL & 55.11           & 60.87          & 62.25          \\
GroundeR + OptiBox + SL & \textbf{58.55}  & \textbf{65.85} & \textbf{67.04} \\
\bottomrule
\end{tabular}
\caption{Comparison against the original GroundeR model in fully-supervised and semi-supervised settings with the semantic loss (SL) with varying proportions of annotations. }
\label{tab: semisupervised}
\end{table}

\textbf{Semi-Supervised:}  Table \ref{tab: semisupervised} also contains results for the semi-supervised learning case. To the best of our knowledge, we are the first to achieve this level of visual grounding accuracy such little annotated data. Comparing to the original semi-supervised GroundeR model, given $3.12\%$ bounding box annotations, our semi-supervised model almost doubled in test accuracy. Moreover, even without OptiBox, the test accuracy of our semi-supervised model ($55.11\%$) performs better than the original GroundeR with full annotations ($48.38\%$). With $50\%$ annotations, there is about $14\%$ increase in test accuracy compared to the original GroundeR with the same amount of annotated data. 

With OptiBox, the accuracy increases further to $65.85\%$, which surpasses most fully-supervised state-of-the-art models (see Table \ref{tab:supervised}) except the concurrent works: GraphGround++ \cite{mohit2019ground} and One-Stage-Bert \cite{yang2019fast}. 

In view of the performance of OptiBox in this semi-supervised setting, when annotated data is scarce ($3.12\%$), the box regression process improves the test accuracy by $3.4\%$. The regressor gives the best improvement ($4.98\%$) when there is $50\%$ annotated data.

\subsection{Ablation Studies: OptiBox}
\begin{figure*}[t]
  \begin{subfigure}[t]{.30\linewidth}
    \centering\includegraphics[width=1\linewidth]{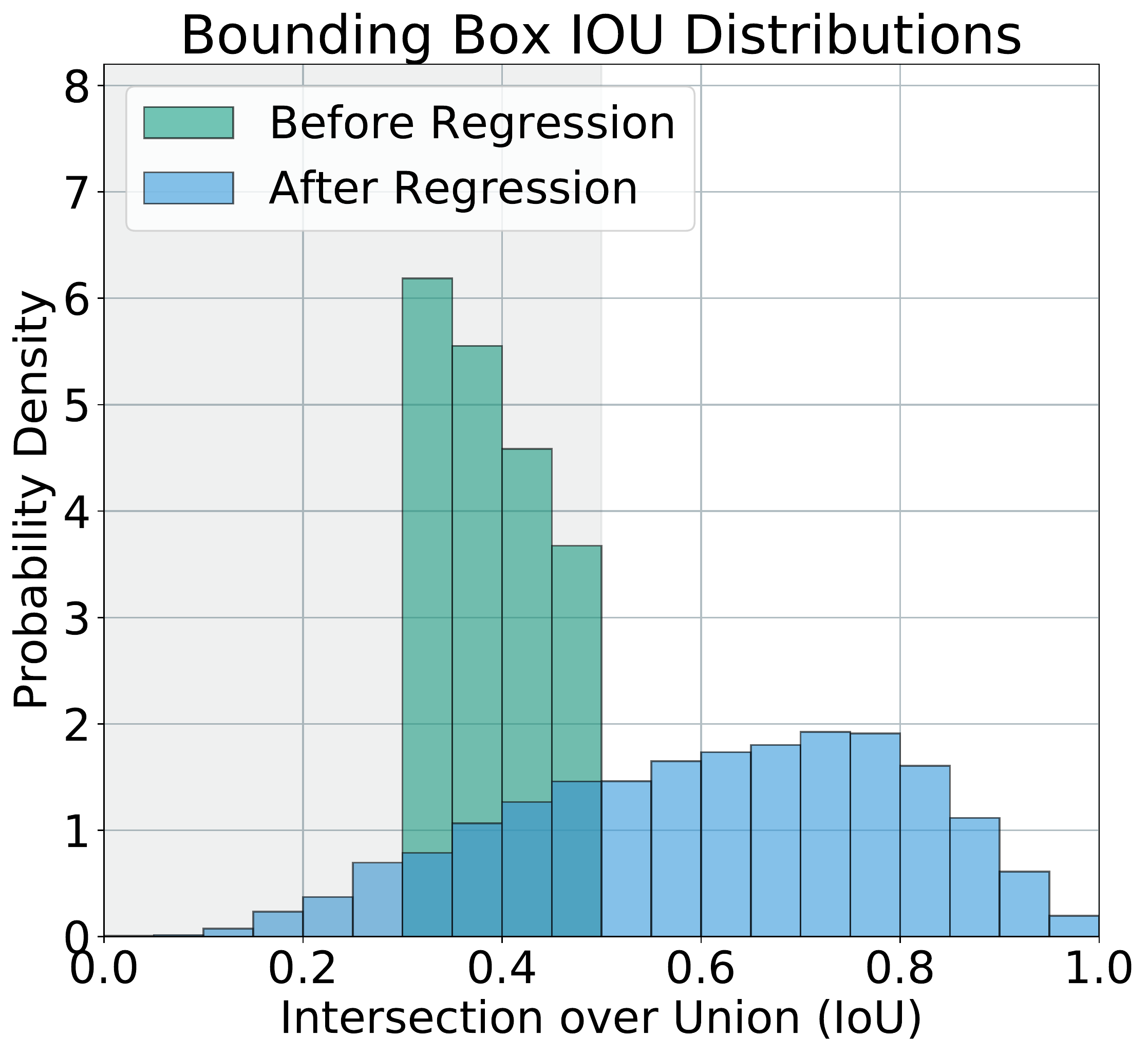}
    \caption{$0.3 \leq \text{IoU} < 0.5$}
  \end{subfigure}\hfill
  \begin{subfigure}[t]{.30\linewidth}
    \centering\includegraphics[width=1\linewidth]{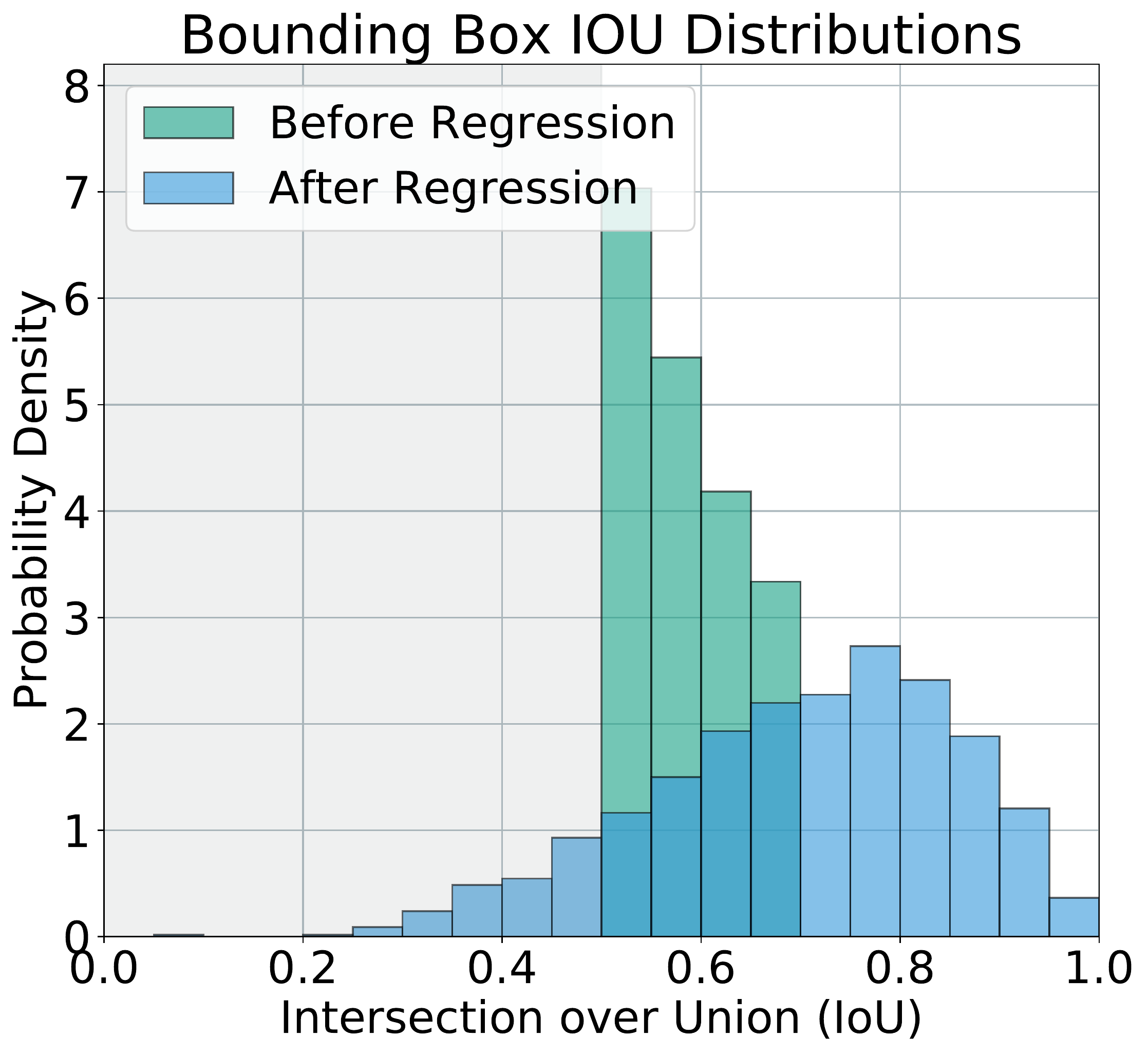}
    \caption{$0.5 \leq \text{IoU} < 0.7$}
  \end{subfigure}\hfill
  \begin{subfigure}[t]{.30\linewidth}
    \centering\includegraphics[width=1\linewidth]{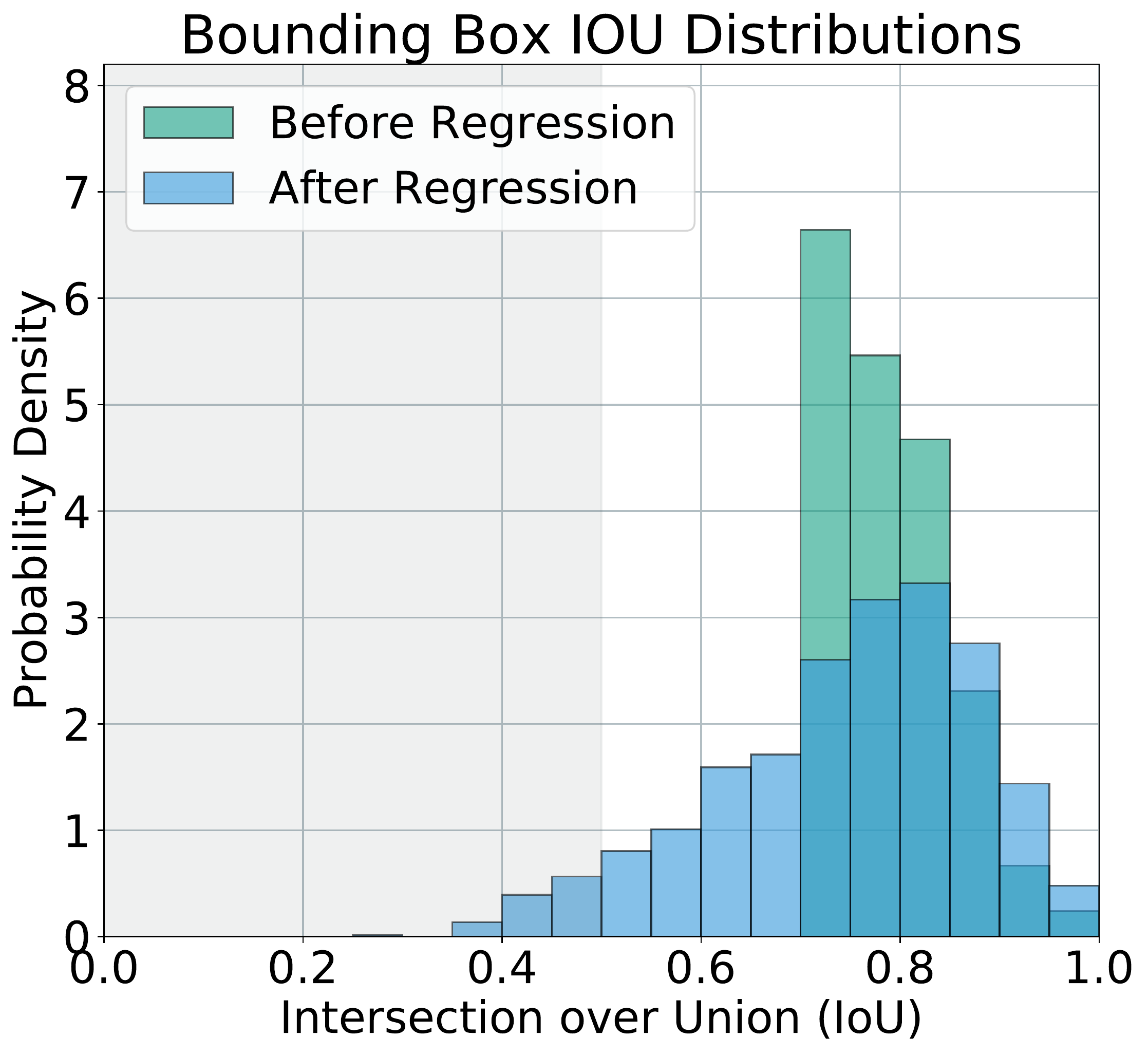}
    \caption{$0.7 \leq \text{IoU} \leq 1.0$}
  \end{subfigure}

  \caption{IoU distributions before and after OptiBox regression. From (a)-(c) we show distribution changes when regressing on different initial qualities of the proposals, in terms of the IoU. Best viewed in color.}
  \label{fig: iou_distribution}
\end{figure*}

In this section, we provide ablation studies to investigate the effect of our bounding box regression model and evaluate how it could be applied to benefit visual grounding models in general.

In Table \ref{tab: regression_ground} we see that the median IoU of model prediction for GroundeR with ResNet101 feature is 0.6008 and 0.6617 when OptiBox is applied, which is $\sim 0.06$ IoU gain on average. Note that if the grounding module gives a negative prediction, it is very likely that the IoU between the chosen box and the ground-truth box is close to zero. Thus, the median IoUs are heavily affected by the zero IoUs. In practice, one would expect the given source bounding box to have at least some overlap with the ground-truth box.

\begin{table}[t]
\centering
\begin{tabular}{ccccc} 
\toprule
\multicolumn{1}{c}{\textbf{Visual}}& \textbf{Box} & \textbf{Query}& \textbf{Global}& \textbf{Median $\Delta$IoU} \\ 
\hline
Y & Y & Y & Y & 0.200 \\
N & Y & Y & Y & 0.106 \\
Y & N & Y & Y & 0.197 \\
Y & Y & N & Y & 0.105 \\
Y & Y & Y & N & 0.198 \\
\bottomrule
\end{tabular}
\caption{Comparison of the effectiveness of the different features for OptiBox. The four feature columns refer to $\mathbf{x}, \mathbf{r}, \mathbf{q}, \mathbf{c}$ as defined in Section \ref{sec: grounder_method}, respectively.
Median $\Delta$IoU is computed by taking the median of IoU differences between bounding box after and before regression.}
\label{tab: features_box_regression}
\end{table}

Since the distribution of IoUs of the predicted boxes is heavily conditioned on the architecture of the grounding model, to assess its universal regression performance, we perform an additional independent experiment for the regression model. Specifically, given a set of Flickr30k images, we select the proposal boxes with greater than $0.3$ IoUs to the ground-truth box to perform regression. The choice of $0.3$ IoUs is to ensure that the proposal boxes $\mathbf{r_i}$ are at least nearby the ground-truth boxes $\mathbf{g_i}$. In short, the dataset consists of source and target pairs of boxes ${(\mathbf{s_i}, \mathbf{g_i})}$ where $IoU(\mathbf{s_i}, \mathbf{g_i}) \geq 0.3$. We use the proposal boxes in the previous Flickr30k experiment as our source boxes and associate each Flickr30k ground-truth box with the proposal boxes that have at least $0.3$ IoU. Using the same data split as the previous experiment, we train our box regression for 40 epochs with a learning rate of $0.001$ and a batch size of $32$ using the Adam optimizer \cite{kingma2014adam}. The learning rate decays by a factor of 0.1 at epoch 3, 10, 20, and 30. 

\begin{figure}[t]
    \centering
    \includegraphics[width=0.49\textwidth]{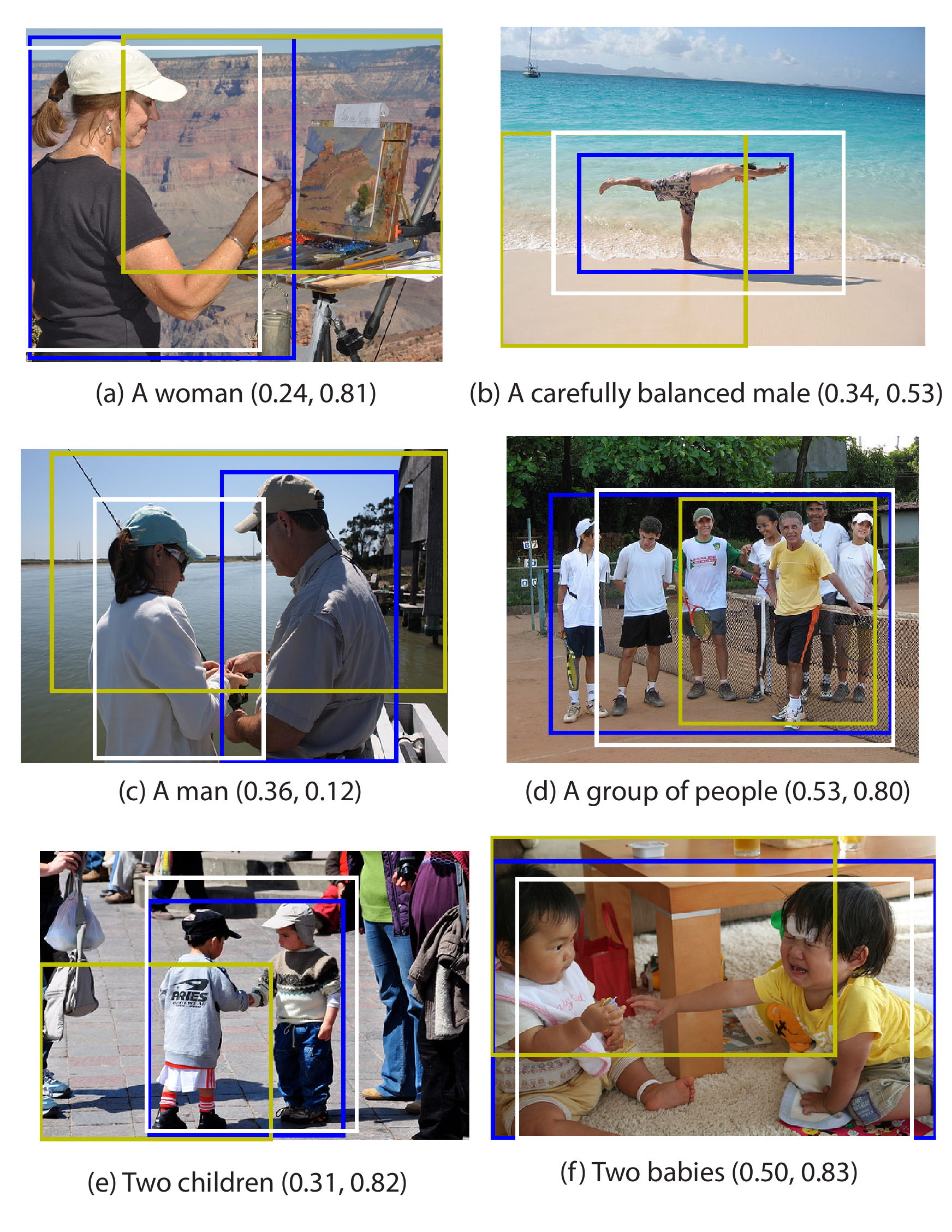}
    \caption{Qualitative samples for OptiBox. Blue boxes are the ground-truths, green boxes are the proposals selected prior to adjustment, and white boxes are the after-adjustment results. Numbers denote the IoU before and after applying OptiBox. Indeed, OptiBox yields more sensible results in most cases. Best viewed in color.}
  \label{fig: qualitative}
\end{figure}

Figure \ref{fig: iou_distribution} shows the distributions of IoUs before and after the bounding box regression in our independent experiment. In particular, Figure \ref{fig: iou_distribution} (a) shows the cases when the given bounding boxes having some but not sufficient overlap ($0.3 \leq \text{IoU} < 0.5$) with the ground truth. We can see that most area of the distribution is moved above 0.5 IoU and only a small area is less than $0.3$ (worse IoU after regression). We denote the region with IoU less than 0.5 by the grey background and denote the white background as IoU great than 0.5. Most source bounding boxes get better off after the regression, and most boxes fall into the region of the white background (IoU greater than 0.5). Looking at Figure \ref{fig: iou_distribution} (b), it shows the cases when the source boxes are relatively good. 
We could see that similar to Figure \ref{fig: iou_distribution} (b), the distribution is moved to a peak of $0.8$ IoU. There is only a tiny area that falls into the grey region. Finally, Figure \ref{fig: iou_distribution} (c) shows the change of IoU distribution for exceptionally good source boxes. Although some portion of source boxes are less than the original IoU minimum ($0.7$), most bounding boxes have IoU $\geq 0.5$, which is the threshold we care about. Figure \ref{fig: qualitative} shows some qualitative examples of our grounding network and our bounding box regression network. For example, Figure \ref{fig: qualitative} (b) shows that the query is ``A carefully balanced male". The bounding box chosen by the grounding network is shown in green. The white box shows the box after the bounding box regression. The blue box is the ground-truth bounding box. The original IoU was 0.34, and it increases to 0.53 after the regression. 
Interestingly, our regression network seems to be flexible with multiple instances. For example, the source bounding box in Figure \ref{fig: qualitative} (e) covers only one child, but the regression network modified it and grounds to two children according to the query. Figure \ref{fig: qualitative} (c) shows a negative example of bounding box regression. Also, differentiating whether a person is ``a man" is often relatively challenging when only their backs are shown in the image.

Finally, Table \ref{tab: features_box_regression} shows the advantages of various features on our regression network. We can see that when all four features are included, our model enjoys a median IoU gain of 0.2 in our independent experiment. The most impactful features are visual features and query features. This makes sense because we expect a box regression model to perform poorly if the model cannot see (\ie, the visual features) or if we do not tell it upon what to focus (\ie, the query). There is a slight decrease in improvement if we drop either the bounding box or the global attention features. This also makes sense because if only a patch on the image is shown, one may still be able to give a decent bounding box prediction of it. However, we would expect global information to be more critical for boxes having less overlap with the ground-truth. In such cases, it is harder to guess bounding boxes if it is unaware of the entire image. Furthermore, the bounding box coordinates provide a slight impact on the regression performance. This is expected because knowing the location and the size of a patch (\ie, the bounding box coordinates) provides some spatial information as to where the attention should be placed.

%% file: sections/conclusion.tex
\section{Conclusion}
In this work, we propose a query-guided box refinement network, OptiBox, which corrects the suboptimal bounding box predicted by a visual grounding model. We demonstrate its effectiveness using the GroundeR model \cite{rohrbach2016grounding} in both supervised and semi-supervised settings. We also introduce a semantic reconstruction loss, which we have shown to provide significant improvement to the overall grounding system. Evaluating on the Flickr30k dataset \cite{plummer2015flickr30k}, we are able to outperform the original fully-supervised GroundeR model with only $3\%$ of the annotations. When we use $50\%$ of the annotations, we are competitive against recently proposed models. In the full-supervised setting, we achieve state-of-the-art performance.

\section*{Acknowledgments}

The authors would like to thank the Natural Sciences and Engineering Research Council of Canada (NSERC) and the Canadian Institute for Advanced Research (CIFAR) for supporting this research project. The authors also appreciate Shih-Han Chou, Bicheng Xu, and Mir Rayat Imtiaz Hossain for helpful feedback and discussions.